\documentclass[pdflatex,sn-mathphys-num]{sn-jnl}

\usepackage{graphicx}%
\usepackage{multirow}%
\usepackage{amsmath,amssymb,amsfonts}%
\usepackage{amsthm}%
\usepackage{mathrsfs}%
\usepackage{xcolor}%
\definecolor{softgray}{gray}{0.95} 
\usepackage{textcomp}%
\usepackage[most]{tcolorbox}
\usepackage[title]{appendix}%
\usepackage{xcolor}%
\usepackage{tikz}
\usetikzlibrary{positioning, arrows.meta}
\usepackage{textcomp}%
\usepackage{manyfoot}%
\usepackage{booktabs}%
\usepackage{caption}  
\usepackage{subcaption}
\usepackage{colortbl}
\usepackage{algorithm}%
\usepackage{algorithmicx}%
\usepackage{algpseudocode}%
\usepackage{listings}%
\tcbset{
  mycssstyle/.style={
    colback=black!5!white,
    colframe=black!75!black,
    listing only,
    listing options={%
      language=HTML,
      basicstyle=\ttfamily\small,
      breaklines=true
    },
    title=CSS Example
  }
}
\theoremstyle{thmstyleone}%

\theoremstyle{thmstyletwo}%

\theoremstyle{thmstylethree}%

\begin{document}

\title[Article Title]{Cycle-of-Thought: Self-Evaluative Iterative Reasoning for Table Visual Question Answering in Document Images}

\author{\fnm{Somraj} \sur{Gautam}}\email{gautam.8@iitj.ac.in}

\author{\fnm{Gaurav} \sur{Harit}}\email{gharit@iitj.ac.in}

\affil{\orgname{IIT Jodhpur}, \orgaddress{\state{Rajasthan}, \country{India}}}

\abstract{Recent advances in Vision-Language Models (VLMs) have significantly improved Visual Question Answering (VQA) performance, yet their reasoning over structured content such as tables remains inconsistent, often plagued by spatial misalignment and logical hallucinations. Existing approaches for improving table VQA typically rely on architectural modifications or expensive supervised fine-tuning (SFT), which impose high barriers to entry due to the need for large-scale, expert-annotated datasets and significant computational resources. In this work, we propose Self-RefineVQA, a training-free, inference-time self-correction framework that improves table VQA performance without additional labeled data or model updates. Inspired by human revision behavior, our method enables a single VLM to iteratively act as a generator, critic, and refiner, forming a closed self-feedback loop that corrects common reasoning errors through prompt-guided re-evaluation. Experiments on tabular VQA benchmarks demonstrate that Self-RefineVQA yields consistent accuracy gains in monolingual and cross-lingual understanding, with a particularly significant boost for small models, enabling lightweight models (e.g., 2B-7B parameters) to rival the performance of significantly larger counterparts. Furthermore, we provide a comparative complexity analysis demonstrating that while inference cost increases linearly, our framework offers superior data efficiency and a lower cumulative cost by eliminating the prohibitive annotation bottleneck of human labeling required for fine-tuning. Our results highlight the potential of prompt-based self-improvement as a scalable, accessible paradigm for reliable table reasoning in VLMs.}


\maketitle

\section{Introduction}\label{sec1}

\begin{figure}[h!]
    \centering
    \begin{subfigure}{0.32\textwidth}
        \centering
        \includegraphics[width=\textwidth]{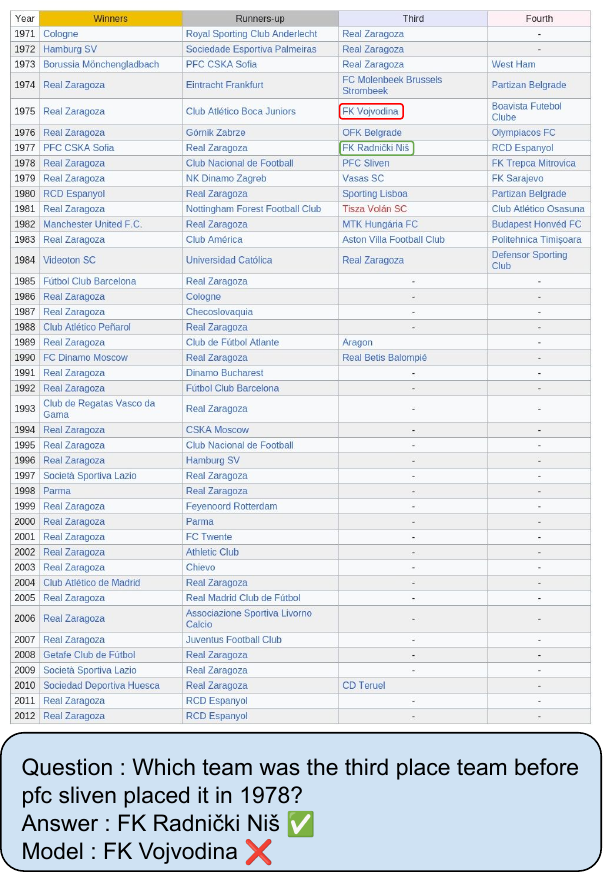}
        \caption{TableVQA-Bench}
    \end{subfigure}
    \hfill
    \begin{subfigure}{0.32\textwidth}
        \centering
        \includegraphics[width=\textwidth]{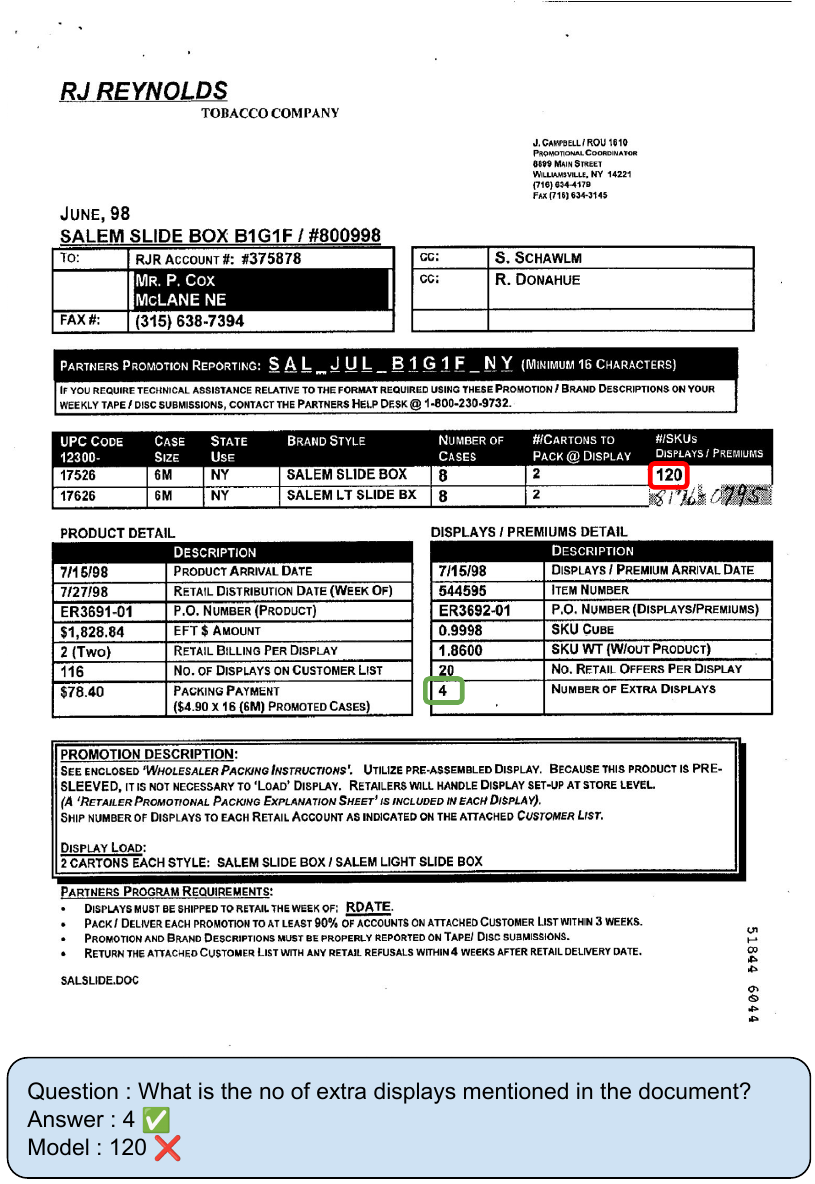}
        \caption{DocVQA-Table}
    \end{subfigure}
    \hfill
    \begin{subfigure}{0.32\textwidth}
        \centering
        \includegraphics[width=\textwidth]{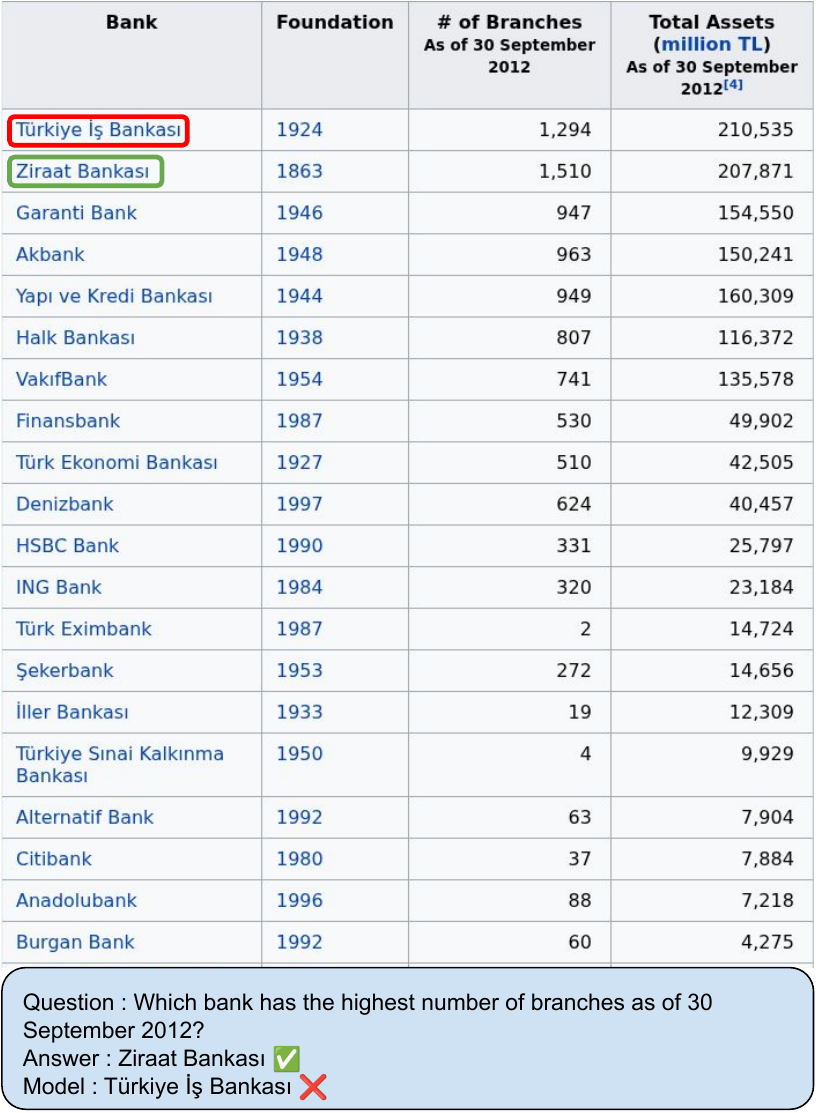}
        \caption{WikiTableQuestions}
    \end{subfigure}
    \caption{\textbf{The Challenge of Spatial Drift in Table VQA.} Examples from (a) TableVQA-Bench, (b) DocVQA, and (c) WTQ. These samples illustrate how minor shifts in attention (spatial drift) lead to reasoning failures, highlighting the need for iterative re-evaluation provided by Self-RefineVQA.}
    \label{fig:intro}
\end{figure}

\begin{figure}[t]
    \centering
    \includegraphics[width=\textwidth]{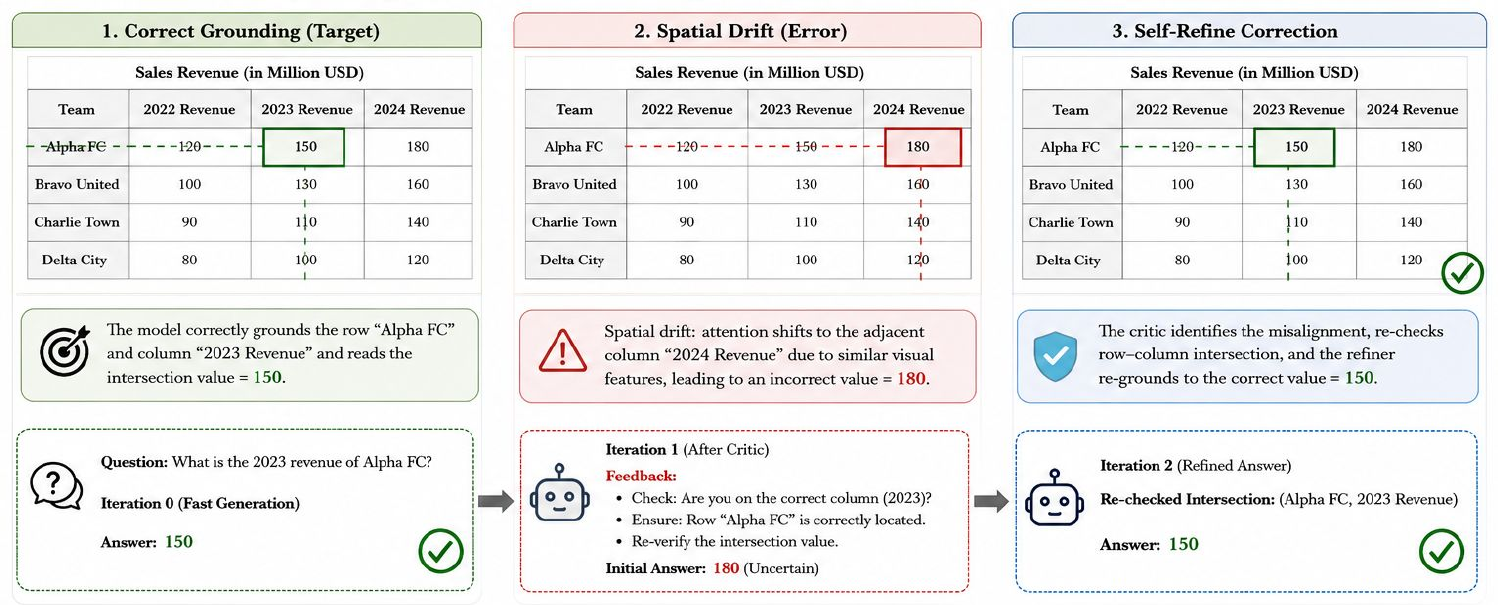}
    \caption{Minor shifts in visual attention cause the model to attend to semantically adjacent but incorrect cells. The proposed Generator–Critic–Refiner cycle detects row-column misalignment and iteratively re-grounds attention to the correct table intersection.}
    \label{fig2}
\end{figure}

Vision–Language Models (VLMs) have rapidly advanced in their ability to parse images, follow natural-language instructions, and perform complex reasoning. Recent large-scale models now achieve impressive results on general Visual Question Answering (VQA), even in zero-shot settings~\cite{li2023blip2, alayrac2022flamingo, liu2023internvl, yang2022unitab}. However, when questions require structured reasoning, such as reading, interpreting, and manipulating information in tables, state-of-the-art VLMs remain unreliable. Table VQA demands multi-step inference across spatial layouts, textual cells, implicit schemas, and arithmetic operations~\cite{zhong2020image,zhong2022pubtabnet,mathew2021docvqa}, while a minor shift in attention in a natural image (e.g., looking slightly to the left of a dog) usually results in a semantically similar representation, a similar shift in a table, from the "2023 Revenue" column to the "2024 Revenue" column—results in a catastrophic error in reasoning, despite the visual features being nearly identical. We term this phenomenon spatial drift, referring to the mismatch between continuous visual embeddings and the discrete symbolic coordinates inherent to tabular data (Figure~\ref{fig:intro}). Spatial drift causes attention to deviate toward visually similar neighboring cells, resulting in incorrect row-column grounding (Figure~\ref{fig2}). Consequently, existing VLMs frequently produce partially correct, inconsistent, or hallucinated answers despite successfully recognizing table structure. This inconsistency is a fundamental obstacle for deploying VLMs in data-intensive domains such as finance, scientific reporting, and business analytics, where correctness is critical.

While self-refinement and critic–revise paradigms have shown success in text-only reasoning tasks, their direct applicability to visual table understanding remains limited. Unlike purely textual inputs, table images introduce partial observability, where errors may originate from misaligned visual attention, incorrect row–column grounding, or perceptual ambiguity rather than faulty logical inference alone.

The prevailing challenge in Table VQA is not necessarily a deficit of 
knowledge or OCR capability, but a deficit of precise execution and 
error detection. When a VLM answers a query, it must execute a precise 
sequence: locating row headers, identifying column intersections, and 
transcribing content. Standard greedy decoding generates answers through sequential token prediction without explicit verification. While efficient, this forward-only process frequently produces errors in table VQA:
\begin{itemize}
    \item Spatial drift: attending to adjacent cells with similar visual features.
    \item Logical inconsistency: inventing constraints not stated in the question.
    \item Premature commitment: selecting plausible but incorrect values.
\end{itemize}These errors often arise not from knowledge gaps but from the absence of a verification phase. Current methodologies to address this fragility typically rely on architectural modifications (e.g., specialized table encoders~\cite{turl2020, yin2020tabert, herzig2020tapas}) or large-scale Supervised Fine-Tuning (SFT)~\cite{wei2021finetuned}. While effective, these approaches impose high barriers to entry: architectural changes break the generality of foundation models, while SFT is data-hungry, requiring tens of thousands of expert-annotated samples, which is an Annotation Bottleneck that makes adaptation computationally expensive and operationally slow.

In this work, we propose Self-RefineVQA, a training-free, model-agnostic, and single-VLM iterative refinement framework designed specifically for table VQA. Unlike prior self-refinement methods that assume fully observed symbolic inputs, our approach explicitly addresses the unique challenges of visual table understanding, including spatial grounding failures, perceptual ambiguity, and incorrect row-column alignment. Our key insight is that modern VLMs implicitly possess multiple abilities: generation, critique, and revision, which can be orchestrated purely through prompting. The Critic prompt guides the model to reverify visual evidence against its initial outputs, enabling the systematic correction of visual-reasoning errors without requiring gradient updates or architectural modifications. 

A central contribution of this work is a rigorous re-evaluation of efficiency. 
We provide a complexity analysis showing that while Self-RefineVQA incurs a marginal increase in inference cost, it bypasses the substantial fixed costs associated with data labeling for model training. This makes high-performance table reasoning accessible for closed-source models and small open-weight models (e.g., 2B–7B parameters), which we demonstrate disproportionately benefit from.

We operationalize this through a three-stage cycle:
\begin{itemize}
    \item \textbf{Generator:} VLM proposes an initial answer;
    \item \textbf{Critic:} Same VLM evaluates the answer via structured critique prompts;
    \item \textbf{Refiner:} Same VLM incorporates the critique to revise the answer.
\end{itemize}
This cycle-of-thought repeats until a stopping criterion or maximum iteration is met, enabling the model to iteratively converge toward more accurate and consistent responses.

Empirically, we evaluate Self-RefineVQA across diverse table VQA datasets and modern VLMs. The method consistently yields step-wise improvements over the model’s zero-shot outputs, surpassing standard self-consistency baselines and approaching (or exceeding) the performance of far more expensive supervised methods. 


Our contributions are summarized as follows:
\begin{itemize}
    \item \textbf{Self-RefineVQA:} A novel, training-free framework that leverages the "Generator-Critic-Refiner" cycle to correct spatial and logical errors in Table VQA without gradient updates.
    \item \textbf{The "Small Model" Boost:} Empirical evidence demonstrating that iterative refinement serves as an equalizer, allowing lightweight models (e.g., Qwen2.5-VL 3B) to achieve gains of up to 13\% on benchmarks like WikiTableQuestions, competing with much larger architectures.
    \item \textbf{Data-Efficient Complexity Analysis:} A formal comparison demonstrating that for most real-world deployments, the data efficiency of self-refinement outweighs the compute efficiency of fine-tuning.
\end{itemize}

This work suggests a paradigm shift from training-centric improvements to inference-centric reasoning, offering a scalable path to reliable document-understanding agents.

\begin{tcolorbox}[colback=softgray,colframe=black,boxrule=0.5pt,arc=0pt]
\centering
\textit{Better answers via prompting, not training}
\end{tcolorbox}

\section{Related Work}
\subsection{Table Understanding and Table VQA}
Early work on table understanding focused on structural parsing, table detection, and cell extraction, forming the foundation for downstream reasoning tasks. Classical methods relied on rule-based extraction and graph-based structure modeling~\cite{camelot2018,tabula,graphtsr2022}, while recent approaches employ deep architectures for table structure recognition and semantic parsing~\cite{tableNet2019, lgpma2022, pubtabnet2020}. Table VQA, in particular, requires combining visual perception with structured reasoning over cell-level representations. Models such as TAPAS~\cite{tapas2020}, TaBERT~\cite{yin2020tabert}, TURL~\cite{turl2020}, and TaPas-based table question-answering systems achieve strong performance but operate primarily on textual or structured table formats rather than raw images. More recent datasets, including WTQ~\cite{wtq2017}, DocVQA-Table~\cite{mathew2021docvqa}, and TableVQA-Bench~\cite{kim2024tablevqa}, encourage end-to-end table reasoning directly from images, revealing that general-purpose VLMs still struggle with layout-sensitive and arithmetic queries.

Image-based table VQA using VLMs remains challenging because models must jointly solve OCR, spatial layout modeling, semantic grounding, and multi-step reasoning. Although OCR-free VLMs, such as LayoutLMv3~\cite{layoutlmv3}, Pix2Struct~\cite{pix2struct2023}, PaLI-X~\cite{paliX2023}, LLaVA~\cite{llava2023}, Llama~\cite{meta2024llama}, Flamingo~\cite{flamingo2022}, InternVL~\cite{internvl2024}, Qwen2-VL~\cite{wang2024qwen2}, Qwen2.5-VL~\cite{bai2025qwen2}, and Qwen3-VL~\cite{Qwen-VL} show improvements, they often hallucinate or provide partially correct answers. While TAPAS is designed for question answering over pre-extracted text tables, it is commonly applied to table images via OCR-based table extraction pipelines, inheriting additional noise from imperfect visual-to-text conversion. Our work improves these models' performance through inference-time refinement rather than architecture or training modifications.

\subsection{Vision–Language Models and Multimodal Reasoning}
Large VLMs integrate visual encoders with LLM backbones, enabling unified reasoning across modalities. Open-source systems (e.g., LLaVA, MiniGPT-4, Qwen-VL, InternVL) and proprietary models (e.g., GPT-4V, Gemini) demonstrate significant progress on perception-rich tasks. However, despite strong zero-shot generalization, their reasoning on structured visual artifacts remains brittle. Several works attempt to improve multimodal reasoning via chain-of-thought prompting, visual grounding modules, retrieval augmentation, or external symbolic tools.

Our approach differs in that we do not modify the model architecture, incorporate external modules, or perform additional training. Instead, we guide a single VLM to critique and revise its own answers, allowing it to leverage latent multimodal reasoning capabilities without supervision.

\subsubsection{Self-Refinement, Critic-and-Revise, and Iterative Reasoning}
Self-refinement has emerged as an effective inference-time strategy for improving language model reasoning, particularly in text-only settings. Methods such as Self-Refine~\cite{madaan2023selfrefine}, Reflexion~\cite{shinn2023reflexion}, and related critic–revise frameworks iteratively evaluate and modify generated answers to reduce logical or factual errors. However, these approaches assume fully observed symbolic inputs, where the primary sources of error lie in reasoning rather than perception. In contrast, table VQA over document images introduces additional challenges, including partial observability, spatial drift, and incorrect row–column grounding. \textbf{These failure modes cannot be resolved through linguistic critique alone.} Our work extends self-refinement to this underexplored setting by introducing schema-aware visual critique, explicitly guiding the model to re-ground its answers in the table’s visual structure during refinement.

It is important to distinguish our contribution from prior self-refinement work at the level of failure mode, not merely at the level of pipeline structure. Methods such as Self-Refine~\cite{madaan2023selfrefine} and Reflexion~\cite{shinn2023reflexion} operate over fully observed symbolic inputs, where the dominant error source is logical inconsistency or factual hallucination in the language space. In contrast, table VQA over document images introduces perceptual errors that cannot be resolved through linguistic critique alone: a model may attend to the correct row label but drift to an adjacent column due to overlapping visual embeddings, producing a confident yet spatially incorrect answer. Our Critic prompt is specifically designed to address this by enforcing explicit row alignment, column alignment, and intersection verification as structured checklist items, which is a form of visual re-grounding that has no direct analog in text-only self-refinement frameworks. The improvement brought by our method is therefore not attributable to the iterative structure per se, but to the structured visual verification that conditions each refinement step on verified spatial coordinates in the image.

\subsubsection{Inference-Time Methods vs. Fine-Tuning}
A parallel line of work attempts to improve VLM reasoning through fine-tuning or full supervised training. Instruction tuning~\cite{llava2023,dai2023instructblip}, LoRA-based adaptation~\cite{hu2022lora}, multimodal alignment, and reinforcement learning~\cite{sun2024aligning} have been highly successful at enhancing task-specific performance. Yet these approaches require labeled data, significant GPU resources, and access to model weights—constraints incompatible with many closed-source or commercial VLMs~\cite{achiam2023gpt, team2023gemini}.


To the best of our knowledge, this is the first work to apply a closed single-model self-critique loop specifically designed around visual row–column grounding for table VQA, without auxiliary critics, fine-tuning, or external tools, making it computationally lightweight and broadly applicable.

Existing approaches either require supervision, synthetic data, or parameter updates, rely on external critics or tool-augmented reasoning, or struggle with understanding structured tables and maintaining numerical consistency. We instead introduce a training-free, closed-loop self-refinement pipeline that strengthens table VQA accuracy through VLM-internal critique and iterative correction, bridging VQA reasoning and test-time optimization.



\section{Methodology}\label{sec2}

\subsection{Problem Formulation}
Let an input consist of a table image $I \in \mathbb{R}^{H \times W \times 3}$ and a natural language question $q \in Q$. A Vision-Language Model (VLM) $\mathcal{M}$ produces a textual answer: \[y=\mathcal{M}(I,q)\]
Traditional VQA systems aim to learn a mapping: \[
M^\star = \arg\min_{\mathcal{\mathcal{M}}} \mathbb{E}_{(I, q, y^\ast)} \left[ \mathcal{L}\big(\mathcal{M}(I, q), y^\ast\big) \right]
\]
where $y\ast$ is the ground-truth answer.
\\
We optimize answers at inference time:
\[y^\star = \arg\max_{y \in \mathcal{Y}} P(y \mid I, q)\]
through iterative self-critique and refinement. We frame this process as analogous to the transition from fast, intuitive responses to slow, deliberative verification, a pattern observed in human problem-solving~\cite{kahneman2011fast}. While we make no claims about the model's internal cognitive architecture, our prompting strategy operationally mimics this dual-phase behavior: initial generation prioritizes fluency and speed, while subsequent critique prioritizes accuracy and verification.
\begin{figure}[t]
    \centering
    \includegraphics[width=\textwidth]{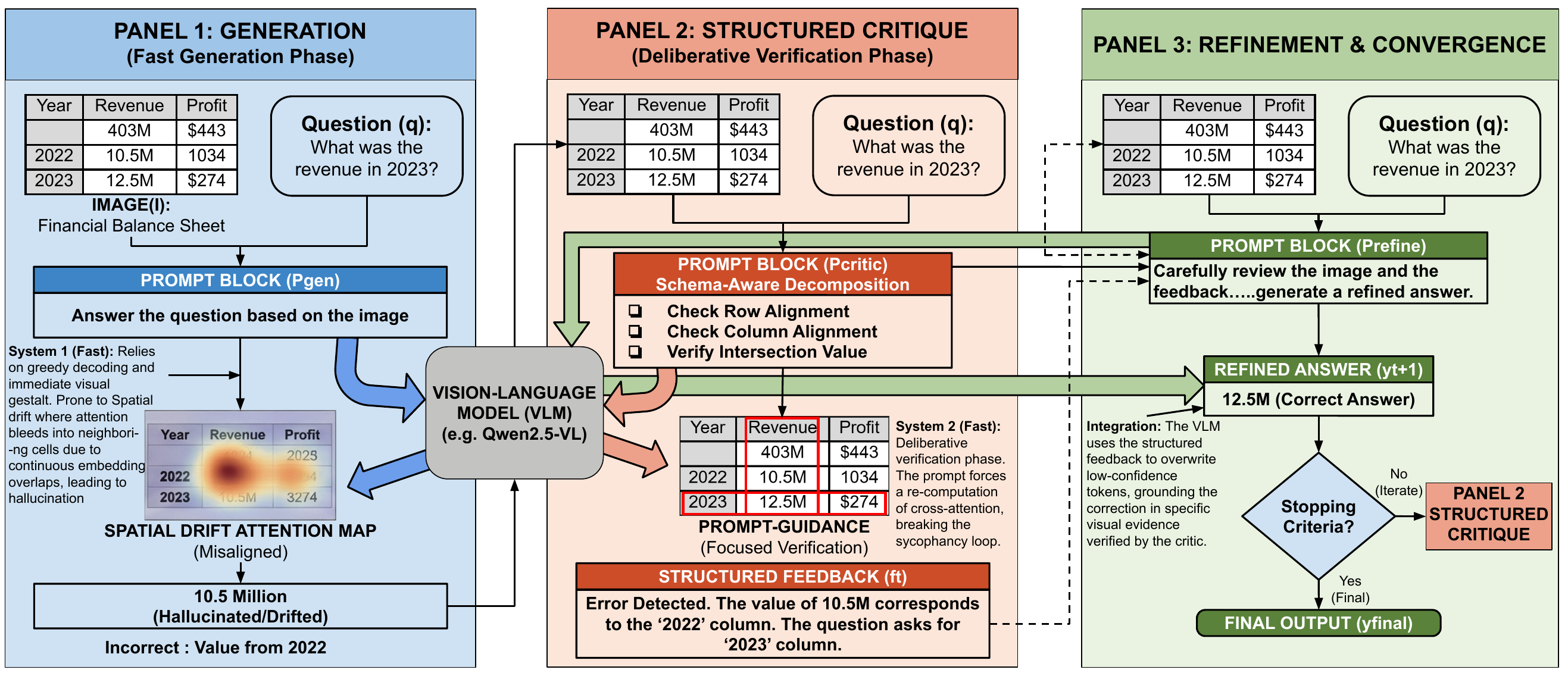}
    \caption{\textbf{The Cycle-of-Thought Framework}: The structured Critic prompt changes the model's conditioning context, prompting re-examination of spatial evidence rather than reinforcement of the initial output. The structured prompting demonstrably reduces error persistence, consistent with a visual re-grounding effect.}
    \label{fig:main_diag}
\end{figure}
\subsection{The Cycle-of-Thought Framework}
Unlike prior self-refinement methods that operate purely in the language space, our Cycle-of-Thought framework, shown in figure~\ref{fig2}, enforces visual re-grounding at every iteration, ensuring that refinement is conditioned on verified row–column intersections in the input image. We decompose inference into three prompting functions: Generator ($G$), Critic ($C$), Refiner ($R$).

The core conceptual insight driving this framework is prompt-guided re-evaluation. In a standard forward pass, a VLM’s reasoning may be influenced by strong textual priors, leading to error persistence once an incorrect answer is generated. By introducing a distinct Critic stage, we prompt the model to re-evaluate the proposed answer $y_t$ against the image $I$ in a separate verification pass.

We introduce prompt-guided re-evaluation, where distinct prompts at the Critic stage change the model's conditioning context, encouraging re-examination of visual evidence rather than reinforcing initial outputs. While we do not claim to directly manipulate attention mechanisms, our structured prompting demonstrably reduces the persistence of errors. This staged prompting functionally separates generation and verification behaviors at inference time, helping break feedback loops in which models otherwise reinforce their own hallucinations.

\subsubsection{Execution Stages}
\textbf{Stage 1: Initial Answer Generation (fast generation):} The VLM produces an initial answer $y_0$ using standard greedy or beam-search decoding: \[y_0 = G(I, q) = \mathcal{M}([I, q], p_G)\]
where $p_G$ denotes a prompt template. We term this "fast generation" - analogous to human intuitive reasoning, the model produces an answer via standard greedy decoding, which may be prone to \textbf{spatial drift} errors, where the model attends to neighboring cells due to continuous embedding overlaps.\\

\textbf{Stage 2: Structured Self-Critique (Deliberative Verification):} In this ``deliberative verification" phase, inspired by human self-checking behavior. We explicitly prompt the model to verify its previous output $y_t$ against visual evidence. Crucially, we leverage the Asymmetry of Verification principle~\cite{ayer1952language}: verifying a solution (classifying ``Match/No-Match") operates in a lower-entropy state than generating a solution from scratch.

Given the current answer $y_t$, the critic produces feedback $f_t \in \mathcal{F}$:
\[f_t = C(I, q, y_t) = \mathcal{M}([I, q, y_t], p_C)\]
We treat critique as a structured decomposition: $f_t=(e_t,\Delta_t)$\\
where:
\begin{itemize}
    \item $e_t$: error identification statement.
    \item $\Delta_t$: suggested correction(s)
\end{itemize}
\textbf{Stage 3: Answer Refinement:} The Refiner integrates the visual evidence with the feedback $f_t$ to overwrite low-confidence tokens:
\[
y_{t+1} = R(I, q, y_t, f_t) = \mathcal{M}([I, q, y_t, f_t], p_R)
\]
where $p_R$ is the refinement prompt. The prompt explicitly enforces dual attention: "Carefully review the image AND the feedback" to ensure the correction is visually grounded.

\subsubsection{Confidence-Aware Stopping Criterion}
To address the stochastic nature of LLMs, we implement a Soft Stopping Condition rather than relying solely on exact string matching. The iterative cycle terminates at step $t$ if either:
\[y_{t+1} =
\begin{cases}
R(I, q, y_t, f_t), & s_t = 0 \\
y_t, & s_t = 1
\end{cases}\]
\begin{itemize}
    \item The Critic outputs a \textbf{No further changes needed} flag ($s_t$) with high confidence (Confidence Score $\ge 4/5$).
    \item The maximum iteration budget $T_{max}$ is reached ($T_{max}=3$).
\end{itemize}
This prevents \textbf{Over-Correction}, where the model might otherwise refine a correct answer into an incorrect one due to hallucinated constraints.

\subsection{Self-RefineVQA Algorithm}
\begin{algorithm}[h]
\caption{Self-RefineVQA}
\begin{algorithmic}[1]
\State \textbf{Input:} Pre-trained Vision-Language Model $\mathcal{M}$; Image $I$; Question $Q$; Ground-truth answer $A_{gt}$; Maximum iterations $K$
\State \textbf{Output:} Refined answer $y_K$

\vspace{0.5em}
\State $y_0 \gets \mathcal{M}(I, Q, P_{gen})$ \Comment{Stage 1, $P_{\text{gen}}$ is the generation prompt}

\vspace{0.5em}

\vspace{0.5em}
\For{$t = 1$ \textbf{to} $K$}
    \State $f_t \gets \mathcal{M}(I, Q, y_{t-1}, P_{critic})$ \Comment{Stage 2: Generate Feedback}
    \If{$\text{"No further changes"} \in f_t$} 
        \State \Return $y_{t-1}$ \Comment{Stop if critic indicates no changes needed}
    \EndIf
    \State $y_t \gets \mathcal{M}(I, Q, y_{t-1}, f_t, P_{refine})$ \Comment{Stage 3: Refine Answer}
    
\EndFor

\State \Return $y_K$ \Comment{Return final answer after $K$ iterations}
\end{algorithmic}
\end{algorithm}

Where, $M$ is Vision-Language Model, $I$ is Input image, $Q$ is Question, $y_t$ is Answer at iteration $t$, $f_t$ is Feedback at iteration $t$, $P_{gen} (p_G), P_{critic}(p_C), P_{refine}(p_R)$ is Stage-specific prompts, $\approx$ is Flexible string matching.

\noindent \textbf{Complexity:} If inference cost is $O(\tau)$: $O_{Self-Refine}=O(T_{max} \cdot \tau)$

\noindent Self-RefineVQA can be viewed as:
\begin{itemize}
    \item Training-free test-time optimization.
    \item A fixed-point iterative reasoning process.
    \item Implicit gradient-free improvement of answer quality.
\end{itemize}


\section{Experiments}
We evaluate Self-RefineVQA on multiple table VQA benchmarks and across a diverse set of modern vision–language models (VLMs). Our experiments aim to answer the following Research questions (RQ):
\begin{itemize}
    \item \textbf{RQ1.}~Does iterative self-refinement improve zero-shot table VQA accuracy across datasets and models?
    \item \textbf{RQ2.}~How does Self-RefineVQA compare to standard inference-time baselines such as self-consistency or multi-sampling?
    \item \textbf{RQ3.}~What is the computational overhead relative to fine-tuning, and is the method cost-efficient?
\end{itemize}

\subsection{Datasets}
We evaluate on three widely used table VQA benchmarks covering diverse domains, layouts, and reasoning types. The statistics of these datasets are summarized in Table~\ref{stats}

\subsubsection{WikiTableQuestions (WTQ) - Image Variant}
We convert WTQ~\cite{wtq2017} tables into rendered image form to assess reasoning over semi-structured tables. WTQ questions require lookup, comparison, superlatives, and aggregation. The dataset is challenging due to varied table schemas and multi-step reasoning.

\subsubsection{DocVQA - Task 2 (Table Images)}
This benchmark includes table images extracted from DocVQA~\cite{mathew2021docvqa} dataset, containing real document images such as invoices, forms, and reports. It emphasizes layout robustness and handling of OCR noise.

\subsubsection{TableVQA-Bench}
A comprehensive benchmark containing tables from financial reports, scientific articles, spreadsheets, and hand-drawn sources. Questions span categories, including arithmetic, span extraction, date operations, and logical comparison.

In DocVQA, we follow the standard train/validation/test splits. Since our method is training-free and the sizes of WikiTableQuestions and TableVQA-Bench are relatively small, we use only the test sets for evaluation, and fine-tuning is performed only for DocVQA-Table.

\subsubsection{IndotabVQA}
To evaluate cross-lingual table reasoning, we include the IndotabVQA~\cite{gautam2026indotabvqa} dataset. This benchmark tests a model's ability to interpret and reason over structured tabular data across diverse linguistic schemas, specifically targeting Arabic, English, Bahasa, and Hindi. It highlights how well inference-time refinement bridges the gap between continuous visual embeddings and multilingual text alignment, with experimental results showing improvement over the fine-tuned model reported in the original paper~\cite{gautam2026indotabvqa}.

\begin{table}[t]
\centering
\caption{Dataset statistics used for evaluating Self-RefineVQA.}
\begin{tabular}{cccc}
\toprule
\multicolumn{1}{c}{\textbf{Dataset}} & \multicolumn{1}{c}{\textbf{Domain/Source}} & \textbf{\# Images} & \textbf{\# QA Pairs} \\ \midrule
WikiTableQuestions (Image) & Wikipedia tables & 2062 & 2062 \\ 
DocVQA - Table & Real documents & 1500 & 1500 \\ 
TableVQA-Bench & Financial reports, scientific articles & 1758 & 1758 \\
IndotabVQA & Real Documents & 1043 & 4172 \\ \bottomrule
\end{tabular}
\label{stats}
\end{table}
\subsection{Models and Baseline}
We evaluate Self-RefineVQA on the following VLM families:
Qwen2-VL (2B, 7B), LLaVA / LLaVA-NeXT, InternVL2
These models differ in size, visual encoder architecture, instruction tuning quality, and OCR-free capabilities, providing a broad testbed for generality.\\
All models are used purely zero-shot. No gradients or parameter updates are performed.

We compare Self-RefineVQA against the following inference strategies:

\begin{itemize}
    \item \textbf{Zero-shot VLM:} A single forward pass using the model's default answering behavior.
    \item \textbf{Self-Consistency (CoT-SC):} We sample multiple CoT. outputs (3) and select the majority-voted answer.
    \item \textbf{Supervised Fine-Tuning:} We report published scores for fine-tuned table QA models (e.g., TAPAS) in which table images are first converted into text tables using OCR and heuristic structure parsing, after which TAPAS is applied without any architectural modification.
\end{itemize}
Our method is orthogonal to training-based approaches.

\subsection{Implementation Details}
\textbf{Prompts}\\
We use three structured prompts:
\begin{tcolorbox}[mycssstyle, title=Generator Prompt ($P_{gen}$): Concise answer generation]
Question : $\{question\}$\\
Answer the question based on the image.
\end{tcolorbox}
\begin{tcolorbox}[mycssstyle, title=Critique Prompt ($P_{critic}$): Error detection + Explanation template]
Question : $\{question\}$\\
Proposed Answer : $\{proposed\_answer\}$\\
You are an expert evaluator. Your task is to analyze the 'Proposed Answer' for the given 'Question' and 'Image'.\\
1. Carefully compare the answer against the information in the table image.\\
2. Identify any inaccuracies, misinterpretations of the data, or incomplete information by checking:\\
- Row Alignment.\\
- Column Alignment. \\
- Verify Intersection Value.\\
3. Provide specific, actionable feedback on how to correct the answer. If the answer is perfect, state "No further changes."\\
Provide your detailed feedback:
\end{tcolorbox}
\begin{tcolorbox}[mycssstyle, title=Refinement Prompt ($P_{refine}$): Revise answer using critique guidance]
Original Question : $\{question\}$\\
Initial Answer : $\{initial\_answer\}$\\
Feedback for Improvement : $\{feedback\}$\\
Carefully review the image and the feedback provided.\\ Your refined answer should:\\
1. Address all the issues mentioned in the feedback.\\
2. Be accurate according to the information in the image.\\
3. Be clear and concise.\\
Provide your refined answer:
\end{tcolorbox}

All prompts are deterministic and fixed across all models unless ablated.\\
\textbf{Iteration Budget:} The maximum number of refinement iterations is: $T_{max}=3$

\subsection{Evaluation Metric}
For WikiTableQuestions rendered as images, we follow the standard WTQ evaluation protocol. Model predictions are compared against ground-truth answers using normalized exact match accuracy. Both predicted and reference answers are normalized by lowercasing, removing punctuation and articles, and standardizing whitespace.

For numerical answers, we allow equivalence under numeric normalization (e.g., 12, 12.0, 12\% when semantically equivalent). For list or multi-span answers, predictions are considered correct if all gold items are present, irrespective of order. This evaluation is consistent across zero-shot, self-consistency, and Self-RefineVQA settings.

\begin{table}[t]
\centering
\caption{Performance of Models on three different dataset.}
\begin{tabular}{lcc>{\columncolor[gray]{0.9}}cl}
\toprule
\textbf{Model} & \textbf{Zero-shot} & \textbf{Self-consistency} & \textbf{Self-refine (Ours)} & \textbf{Improvement} \\
\midrule
\multicolumn{5}{c}{\cellcolor[gray]{0.9}{\textbf{WikiTableQuestions}}} \\ \midrule
TAPAS               & 24.10 & 24.10 & 26.96 & \textbf{+ 2.86} \\ 
Qwen2-VL [2B]       & 53.54 & 58.58 & 61.06 & \textbf{+ 7.52} \\ 
Qwen2.5-VL [3B]     & 58.97 & 56.45 & 71.97 & \textbf{+ 13.00} \\
Qwen3-VL [2B]       & 65.86 & 44.40 & 69.59 & \textbf{+ 3.73} \\
Qwen2-VL [7B]       & 63.29 & 65.91 & 74.59 & \textbf{+ 11.30} \\
Qwen2.5-VL [7B]     & 63.09 & 58.92 & 72.55 & \textbf{+ 9.46} \\
Qwen3-VL [8B]       & 72.99 & 53.25 & 75.90 & \textbf{+ 2.91} \\
InternVL3 [8B]      & 49.22 & 53.30 & 63.09 & \textbf{+ 13.87} \\
Llama-3.2 [11B]     & 57.37 & 53.44 & 66.05 & \textbf{+ 8.68} \\ \midrule
\multicolumn{5}{c}{\cellcolor[gray]{0.9}{\textbf{DocVQA-Table}}} \\ \midrule
TAPAS              & 35.84 & 35.95 & 38.91 & \textbf{+ 3.07} \\ 
Qwen2-VL [2B]      & 90.59 & 91.98 & 92.07 & \textbf{+ 1.48} \\
Qwen2.5-VL [3B]    & 89.97 & 83.11 & 93.47 & \textbf{+ 3.50} \\
Qwen3-VL [2B]      & 90.28 & 75.31 & 91.91 & \textbf{+ 1.63} \\
Qwen2-VL [7B]      & 93.29 & 96.25 & 94.46 & \textbf{+ 1.17} \\
Qwen2.5-VL [7B]    & 91.76 & 87.26 & 93.31 & \textbf{+ 1.55} \\
Qwen3-VL [8B]      & 93.93 & 73.15 & 95.02 & \textbf{+ 1.09} \\
InternVL3 [8B]     & 86.86 & 75.88 & 91.68 & \textbf{+ 4.82} \\
Llama-3.2 [11B]    & 89.42 & 86.52 & 92.07 & \textbf{+ 2.65} \\ \midrule
\multicolumn{5}{c}{\cellcolor[gray]{0.9}{\textbf{TableVQA-Bench}}} \\ \midrule
TAPAS              & 24.10 & 24.40 & 26.96 & \textbf{+ 2.86} \\
Qwen2-VL [2B]      & 43.20 & 46.60 & 57.20 & \textbf{+ 14.00} \\
Qwen2.5-VL [3B]    & 52.67 & 48.13 & 72.80 & \textbf{+ 20.13} \\
Qwen3-VL [2B]      & 71.27 & 41.00 & 75.53 & \textbf{+ 4.27} \\
Qwen2-VL [7B]      & 48.53 & 56.53 & 66.60 & \textbf{+ 18.07} \\
Qwen2.5-VL [7B]    & 71.53 & 87.26 & 79.87 & \textbf{+ 8.33} \\
Qwen3-VL [8B]      & 82.80 & 54.00 & 86.73 & \textbf{+ 3.93} \\
InternVL3 [8B]     & 58.40 & 47.47 & 70.87 & \textbf{+ 12.47} \\
Llama-3.2 [11B]    & 61.53 & 49.93 & 71.67 & \textbf{+ 10.13} \\
\bottomrule
\label{Result}
\end{tabular}
\end{table}

\section{Results}\label{sec5}
We present the evaluation of Self-RefineVQA across three diverse benchmarks: WikiTableQuestions (WTQ), DocVQA-Table, and TableVQA-Bench. Table~\ref{Result} summarizes the zero-shot performance, self-consistency baselines, and our proposed self-refinement results across varying model scales.

\subsection{Main Results: The Small Model Boost (RQ1)}
Our experiments reveal a consistent performance improvement across all tested models.
\begin{itemize}
    \item \textbf{Significant Gains for Lightweight Models:} The largest improvements are observed in the 2B–3B parameter regime. For instance, on the WikiTableQuestions (WTQ) dataset, Qwen2.5-VL (3B) improves from a baseline of 58.97\% to 71.97\%, a substantial gain of +13.00\%. Similarly, InternVL3 (8B) exhibits a +13.87\% improvement.
    \item \textbf{Diminishing Returns for Larger Models:} As model capacity increases, the relative gain from self-refinement decreases. For example, Qwen3-VL (8B) improves by only +2.91\% on WTQ (72.99\% $\to$ 75.90\%).
    \item \textbf{Cross-Lingual Robustness on IndotabVQA}: Iterative refinement serves as a powerful equalizer for cross-lingual spatial reasoning shown in table~\ref{tab:indotabvqa}. On the IndotabVQA dataset, Qwen2.5-VL (7B) demonstrated the largest improvements, securing a +7.96\% gain in Bahasa and +6.14\% in both Arabic and Hindi. Qwen2-VL (7B) followed closely with consistent gains across all four languages, proving that prompt-guided re-evaluation effectively mitigates spatial drift even when the underlying tabular schema shifts across different languages.
\end{itemize}

\begin{figure}[t]
    \centering
    \begin{subfigure}{0.32\textwidth}
        \centering
        \includegraphics[width=\textwidth]{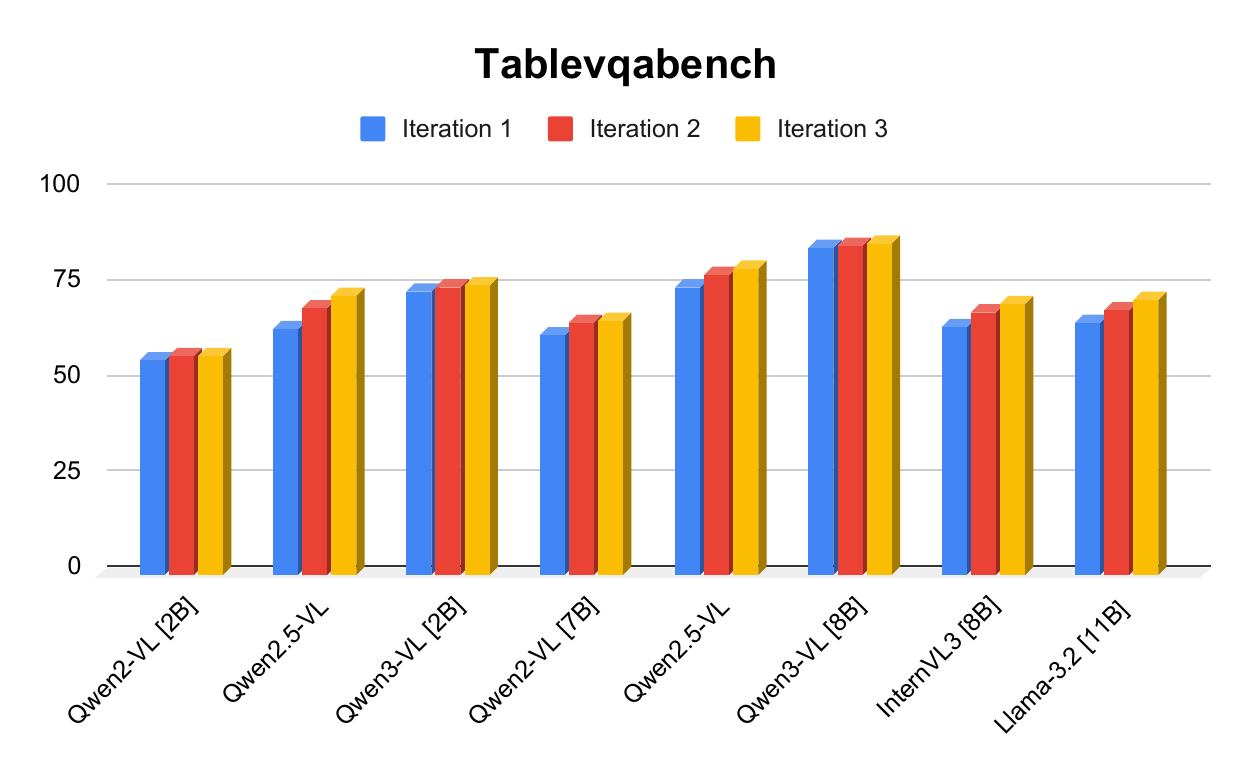}
        \label{fig:tablevqa}
    \end{subfigure}
    \hfill
    \begin{subfigure}{0.32\textwidth}
        \centering
        \includegraphics[width=\textwidth]{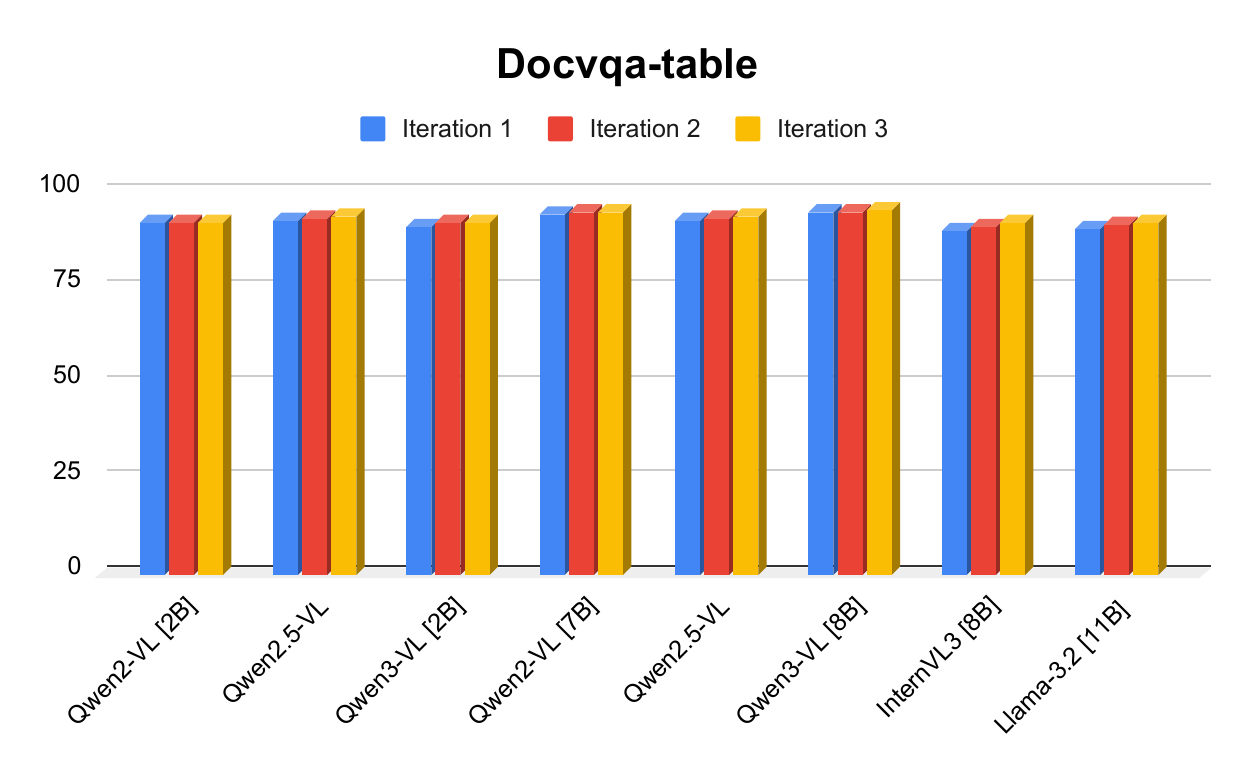}
        \label{fig:docvqa}
    \end{subfigure}
    \hfill
    \begin{subfigure}{0.32\textwidth}
        \centering
        \includegraphics[width=\textwidth]{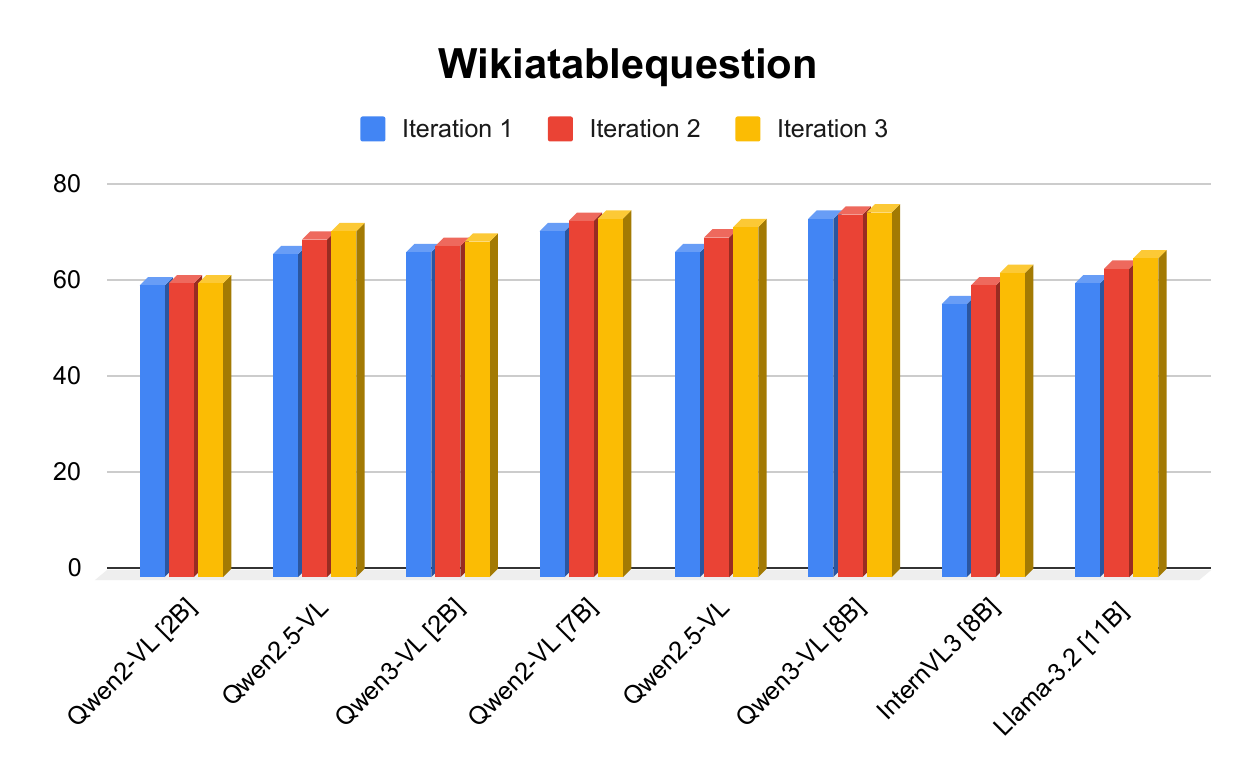}
        \label{fig:wikitable}
    \end{subfigure}
    \caption{Iteration-wise accuracy progression across WikiTableQuestions, DocVQA-Table, and TableVQA-Bench. Most gains occur in the first 1-2 iterations, after which performance saturates, supporting our choice of Tmax = 3.}
    \label{fig:three-pdfs}
\end{figure}

Figure~\ref{fig:three-pdfs} illustrates the iteration-wise accuracy progression of different models across WikiTableQuestions, DocVQA-Table, and TableVQA-Bench. We observe that most performance gains are achieved within the first one to two refinement cycles, after which accuracy typically saturates. This behavior supports our choice of a small iteration budget ($T_{\max}=3$) and indicates stable convergence rather than over-correction.

These results suggest that smaller models often possess the necessary underlying knowledge to answer table-based questions but are more prone to minor errors, such as spatial drift, formatting inconsistencies, or premature termination, which can be effectively corrected through self-critique. In contrast, larger models tend to make fewer recoverable errors; their remaining failures are more often due to fundamental perceptual limitations (e.g., severe occlusion), which iterative critique cannot resolve. 

\begin{table}[t]
\centering
\scriptsize
\setlength{\tabcolsep}{3.5pt}
\renewcommand{\arraystretch}{1.05}
\resizebox{\columnwidth}{!}{%
\begin{tabular}{llccccc}
\toprule
\textbf{Lang.} & \textbf{Model} & \textbf{Zero-shot} &
\textbf{Self-Cons.} & \textbf{Self-refine} &
\textbf{$\Delta$} \\
\midrule

\multirow{2}{*}{Arabic}
& Qwen2-VL      & 26.75 & -- & \textbf{30.49} & +3.74 \\
& Qwen2.5-VL    & 26.94 & -- & \textbf{33.08} & +6.14 \\

\midrule
\multirow{2}{*}{English}
& Qwen2-VL      & 39.41 & -- & \textbf{44.49} & +5.08 \\
& Qwen2.5-VL    & 40.08 & -- & \textbf{45.45} & +5.37 \\

\midrule
\multirow{2}{*}{Bahasa}
& Qwen2-VL      & 69.51 & -- & \textbf{75.26} & +5.75 \\
& Qwen2.5-VL    & 68.84 & -- & \textbf{76.80} & +7.96 \\

\midrule
\multirow{2}{*}{Hindi}
& Qwen2-VL      & 25.50 & -- & \textbf{29.63} & +4.12 \\
& Qwen2.5-VL    & 24.54 & -- & \textbf{30.68} & +6.14 \\

\bottomrule
\end{tabular}%
}
\caption{Performance of Qwen-VL models on IndotabVQA across different languages. 
Self-refine denotes our proposed approach.}
\label{tab:indotabvqa}
\end{table}

\subsection{Comparison with Baselines (RQ2)}
Self-RefineVQA consistently outperforms the Self-Consistency (CoT-SC) baseline. For example, on TableVQA-Bench, Qwen2.5-VL (3B) achieves 48.13\% with Self-Consistency compared to our method, which achieves 72.80\% accuracy, indicating that iterative critique is more efficient than majority voting for structured data.

We observe that self-consistency occasionally degrades performance 
(e.g., Qwen3-VL 8B: 72.99\% $\to$ 53.25\% on WTQ). Analysis reveals this 
occurs when the model confidently produces the same incorrect answer 
across multiple samples (mode collapse), causing majority voting to 
amplify errors rather than correct them. Self-RefineVQA avoids this 
failure mode by explicitly identifying and correcting errors through 
critique rather than relying on answer agreement.

\textbf{The TAPAS Anomaly:} We observe that the specialized table model TAPAS significantly underperforms modern VLMs (e.g., 24.10\% on WTQ vs. 53.54\% for Qwen2-VL 2B) for the WikiTableQuestions dataset. It is crucial to note that this is an architectural comparison rather than a purely reasoning one. TAPAS relies on an external pipeline to convert images to text (OCR) before reasoning. Its low scores likely reflect upstream OCR failures rather than reasoning deficits, confirming the superiority of End-to-End VLMs for visual table understanding.

\subsection{Dataset-Specific Performance}
The efficacy of refinement varies by domain, revealing that self-critique is more effective at correcting logical errors than perceptual failures:
\begin{itemize}
    \item \textbf{High Gains on Logic Tasks (WTQ):} We observe the largest gains on WikiTableQuestions (up to +13\%), a dataset requiring sorting, comparison, and arithmetic. This confirms that the Critic is highly effective at correcting logical inconsistencies.
    \item \textbf{Modest Gains on Perception Tasks (DocVQA):} Gains on DocVQA-Table are consistently lower (e.g., +1.48\% for Qwen2-VL 2B). DocVQA emphasizes extracting information from noisy real-world scans. If the model cannot visually perceive the text due to blur or noise, the Critic cannot "reason" it into existence. This suggests that visual grounding remains the hard ceiling for self-correction.
    \item \textbf{Multilingual Alignment (IndotabVQA)}: When testing on cross-lingual tables, the refinement process yields solid, consistent gains~\ref{tab:indotabvqa}. The models show strong adaptability to linguistic shifts, indicating that the Critic's structured visual grounding helps overcome the perceptual ambiguity introduced by non-English character sets and varied text directionality (e.g., Arabic)
\end{itemize}

\subsection{Qualitative Error Analysis}
\begin{figure}[t]
    \centering
    \includegraphics[width=\textwidth]{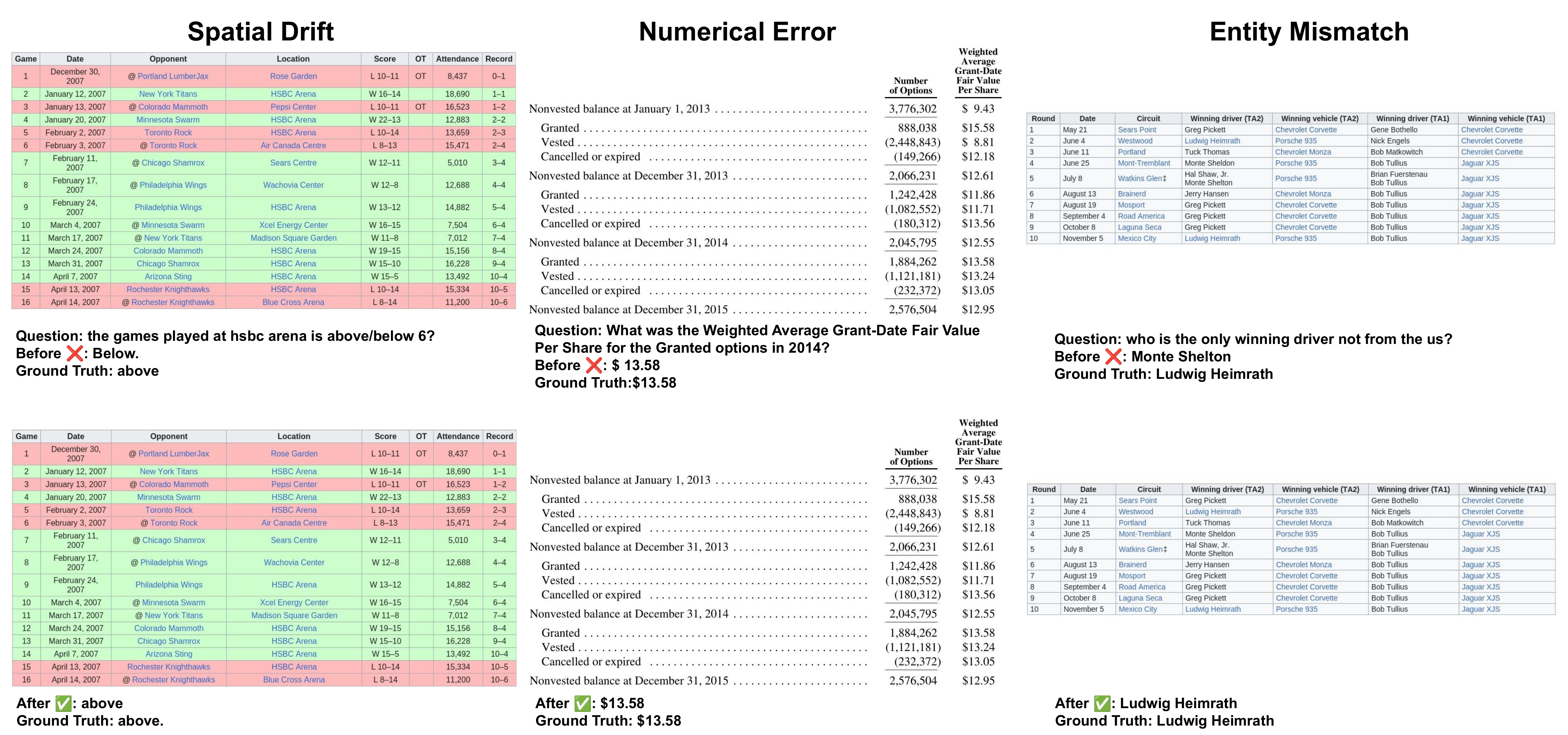}
    \caption{\textbf{Failure taxonomy before and after Self-RefineVQA}: 
We illustrate representative failure modes of zero-shot table VQA models (top), including spatial drift, numerical reasoning errors, and entity mismatches. Self-RefineVQA corrects these errors via iterative self-critique and refinement (bottom), re-grounding predictions in verified row–column intersections without additional training.}
    \label{Error}
\end{figure}

Figure~\ref{Error} provides representative qualitative examples illustrating how Self-RefineVQA corrects common table reasoning errors that arise from spatial drift, numerical inconsistency, and entity mismatch.

\subsection{Ablation Study}
To understand the contribution of each stage in the Cycle-of-Thought framework, we conduct ablation experiments by selectively removing individual stages. We evaluate two reduced configurations against the full three-stage pipeline:
\begin{itemize}
    \item Stage 1 + Stage 2 only (Gen + Critique, no Refine): The model generates an initial answer and produces structured critique feedback, but no refinement pass is performed. The critique output is examined but the original answer is returned unchanged. This isolates whether the Critic stage alone, without an explicit correction step, is sufficient.
    \item Stage 1 + Stage 3 only (Gen + Refine, no Critique): The model generates an initial answer and then directly performs a refinement pass, but without any structured critique feedback. The Refiner prompt receives only the question and the initial answer, without the structured error decomposition. This tests whether unguided re-reading of the image is sufficient for improvement.
\end{itemize}
Results are reported in Table~\ref{ablation}. The Stage (1+2) configuration (Critique without Refinement) produces no change over the zero-shot baseline across all models and datasets. The final accuracy equals the initial accuracy in every case. This is expected by design: the Critic stage identifies and articulates errors but has no mechanism to act on them; without the Refiner, the original answer is returned unchanged. This result cleanly confirms that the Critic stage alone is insufficient and that the Refiner is the necessary executor of corrections.

The Stage (1+3) configuration (Refinement without structured Critique) yields modest improvements; for example, Qwen2.5-VL (7B) improves from 63.48\% to 68.57\% on WTQ and from 72.40\% to 75.53\% on TableVQA-Bench, suggesting that a second pass over the image has some intrinsic benefit. However, these gains fall consistently short of the full pipeline, with gaps of up to 4–5 percentage points on WTQ. This indicates that unguided re-reading of the image lacks the precision of structured error identification: the model improves somewhat by re-attending to the image, but without an explicit diagnosis of which row–column alignment failed, the correction is unreliable. The full pipeline, which combines structured error identification with a targeted refinement pass, achieves the highest accuracy across all settings, confirming that both stages contribute independently and that their interaction is what drives the reported gains.

\begin{table}[t]
\centering
\caption{Ablation study.}
\label{tab:ablation}
\small
\setlength{\tabcolsep}{1pt}
\renewcommand{\arraystretch}{1.1}

\begin{tabular}{llcccc}
\toprule
Dataset & Model &
\shortstack{Stage 1 Only\\(Zero-shot)} &
\shortstack{Stage (1+2)\\(No Refine)} &
\shortstack{Stage (1+3)\\(No Critique)} &
\shortstack{Full Pipeline\\(Ours)} \\
\midrule
WikiTableQuestions & Qwen2.5-VL [7B] & 63.48 & 63.48 & 68.57 & 72.55 \\
                   & Qwen3-VL [8B]   & 73.62 & 73.62 & 74.35 & 75.90 \\
\midrule
DocVQA-Table       & Qwen2.5-VL [7B] & 89.48 & 89.48 & 90.27 & 93.31 \\
                   & Qwen3-VL [8B]   & 93.63 & 93.63 & 94.37 & 95.02 \\
\midrule
TableVQA-Bench     & Qwen2.5-VL [7B] & 72.40 & 72.40 & 75.53 & 79.87 \\
                   & Qwen3-VL [8B]   & 82.07 & 82.07 & 84.07 & 86.73 \\
\bottomrule
\end{tabular}
\end{table}

\section{Complexity Analysis}

Previous works often compare methods solely on inference latency. However, this view ignores the substantial fixed initialization overhead associated with supervised methods. In this section, we provide a formal analysis of the Cumulative Cost for Table VQA deployment, demonstrating that Self-RefineVQA offers a superior economic profile by trading marginal inference compute for massive data efficiency.

\subsection{Formal Cost Formulation}\label{sec6.1}
We define the total lifecycle cost $\mathcal{C}$ of a VQA system as the sum of data acquisition, training compute, and deployment inference. Let:
\begin{itemize}
    \item $N_{train}$: Number of expert-annotated training samples required for Supervised Fine-Tuning (SFT).
    \item $C_{data}$: Cost to acquire and label one sample (human expert effort).
    \item $C_{train}$: Compute cost to train on one sample (Forward + Backward pass).
    \item $N_{test}$: Number of inference queries in the deployment lifetime.
    \item $C_{inf}$: Compute cost for one inference pass (Forward pass).
    \item $K$: Average number of refinement iterations (where typically $K \approx 1.5$).
\end{itemize}
\textbf{Cost of Supervised Fine-Tuning (SFT):} The cost of SFT is dominated by the "Annotation Bottleneck", the fixed cost of creating a high-quality dataset.
$$\mathcal{C}_{SFT} = \underbrace{(N_{train} \cdot C_{data})}_{\text{Annotation Bottleneck}} + \underbrace{(E \cdot N_{train} \cdot C_{train})}_{\text{Optimization Overhead}} + \underbrace{(N_{test} \cdot C_{inf})}_{\text{Marginal Inference}}$$
where $E$ is the number of epochs.\\
\textbf{Cost of Self-RefineVQA (Ours):} Our method incurs zero setup cost. The cost is purely marginal.
$$\mathcal{C}_{SR} = 0 + 0 + \underbrace{(N_{test} \cdot (1 + 2K) \cdot C_{inf})}_{\text{Iterative Inference Cost}}$$
The term $(1+2K)$ accounts for the initial generation plus $K$ cycles of critique and refinement.

\subsection{The Critical Deployment Threshold and Annotation Bottleneck (RQ3)}
Self-RefineVQA is economically superior whenever $\mathcal{C}_{SR} < \mathcal{C}_{SFT}$. Substituting the cost expressions from Sections~\ref{sec6.1}:
$$N_{test}(1+2K) \cdot C_{inf} < N_{train}C_{data} + E \cdot N_  {train}C_{train} + N_{test} \cdot C_{inf}$$

With $K$=1.5, the multiplier $(1+2K)$=4, giving:
$$4 \cdot N_{test} \cdot C_{inf} < N_{train}C_{data} + E \cdot N_  {train}C_{train} + N_{test} \cdot C_{inf}$$
Subtracting $N_{test} \cdot C_{inf}$ from both sides:
$$3 \cdot N_{test} \cdot C_{inf} < N_{train}(C_{data} + E \cdot C_{train})$$
Rearranging for the deployment volume $N_{test}$: $$N_{test} < \frac{N_{train}(C_{data}+E\cdot C_{train})}{3 \cdot C_{inf}}$$

This is the general threshold, which grows with annotation cost $C_{data}$. Since expert annotation of tabular data is expensive and slow, $C_{data} \gg C_{inf}$ in practice, making this threshold extremely large and broadly favorable to Self-RefineVQA.

Conservative Lower Bound Analysis. To establish a worst-case guarantee, we set $C_{data} = 0$ and $E = 1$, which concedes annotation cost entirely to the SFT baseline:
$$N_{test} < \frac{N_{train}\cdot C_{train}}{3 \cdot C_{inf}}$$

Since fine-tuning requires both a forward and backward pass, $C_{train} \approx 2-3 \times C_{inf}$, which gives:

$$N_{test} \lesssim \frac{2}{3} N_{train}$$

This reveals a critical insight: even when annotation is treated as free, Self-RefineVQA remains economically superior as long as the number of deployment queries does not greatly exceed the size of the training set (a condition that holds in the majority of real-world deployments).

Crucially, this is a conservative lower bound. In practice, $C_{data} > 0$ and often $C_{data} \gg C_{train}$, meaning the true threshold is substantially larger, further extending the deployment range over which Self-RefineVQA is the more economical choice.

\subsection{Operational Comparison}
Beyond cost, Self-RefineVQA offers distinct operational advantages in memory and adaptability, as summarized in Table~\ref{table2}.
\begin{table}[h]
\centering
\caption{Operational Efficiency Comparison}
\label{table2}
\begin{tabular}{lcc}
\toprule
\textbf{Metric} & \textbf{Supervised Fine-Tuning (SFT)} & \textbf{Self-RefineVQA (Ours)} \\
\midrule
Data & High (Requires 1k+ labeled pairs) & Zero (Training-free) \\
Compute & High (Backprop, Optimizer states) & Zero (Inference only) \\
Inference Cost & Low ($1\times T_{\text{gen}}$) & Moderate ($\approx 4\times T_{\text{gen}}$) \\
Memory & High (Weights + Gradients + Optimizer) & Low (Weights only) \\
Adaptability & Low (Requires retraining for new schemas) & High (Prompt adjustment) \\
Access & Open Weights (Required) & Black Box API (Compatible) \\
\bottomrule
\end{tabular}
\end{table}

\subsection{Memory Efficiency}
Fine-tuning requires storing gradients, optimizer states (e.g., Adam), and activations, often demanding 4$\times$ to 6$\times$ the VRAM of inference. A 7B model capable of inference on consumer hardware (e.g., 24GB VRAM) may require A100-class hardware (80GB) for fine-tuning. Self-RefineVQA operates strictly within inference memory budgets, democratizing access to high-performance reasoning for researchers with limited hardware.

\section{Discussion and Broader Impact}
The success of Self-RefineVQA challenges the prevailing assumption in the Document AI field that higher performance requires increasingly specialized architectures or massive task-specific fine-tuning. Instead, we demonstrate that latent reasoning capabilities already exist within strong foundation models but require structured inference-time scaffolding to be unlocked.

By eliminating the need for fine-tuning, our framework democratizes access to high-performance table reasoning. Small organizations and researchers who cannot afford to train 70B parameter models or curate 100k+ sample datasets can achieve state-of-the-art results using off-the-shelf models and our prompt pipeline. This has significant implications for privacy-sensitive domains (e.g., healthcare, legal) where data cannot be sent to third-party training services.

Future Directions: Future work should explore Multi-Modal Retrieval Augmented Generation (RAG), integrating this iterative refinement with retrieval systems for multi-page documents. Additionally, Reinforcement Learning for Prompt Optimization (RLPO) could be employed to automatically discover the optimal Critic prompts, replacing our heuristic templates with learned instructions that maximize the "Correct-to-Correct" transition probability.
\section{Conclusion}
In this work, we presented Self-RefineVQA, a training-free framework that enables Vision-Language Models to iteratively correct their own reasoning errors on tabular data. By structuring inference as a Cycle-of-Thought (Generation, Critique, and Refinement), we achieved state-of-the-art results on WikiTableQuestions, DocVQA-Table, and IndotabVQA using off-the-shelf models in both monolingual and cross-lingual table VQA capabilities. Our extensive empirical evaluation demonstrates that the bottleneck in Table VQA is often not the model's underlying knowledge, but its attention management and logical execution, which can be significantly improved through iterative self-evaluation. This approach offers a scalable, privacy-preserving, and computationally efficient path toward reliable document understanding agents.


\bibliography{sn-bibliography}

@article{li2023blip2,
  title={BLIP-2: Bootstrapping Language-Image Pre-training with Frozen Image Encoders and Large Language Models},
  author={Li, Junnan and Li, Dongxu and Savarese, Silvio and Malik, Jitendra},
  journal={arXiv preprint arXiv:2301.12597},
  year={2023}
}

@inproceedings{alayrac2022flamingo,
  title={Flamingo: a Visual Language Model for Few-Shot Learning},
  author={Alayrac, Jean-Baptiste and others},
  booktitle={NeurIPS},
  year={2022}
}

@inproceedings{liu2023internvl,
  title={InternVL: Scaling up Vision Foundation Models and Aligning for Generic Visual-Linguistic Tasks},
  author={Liu, Hao and others},
  booktitle={CVPR},
  year={2024}
}

@article{yang2022unitab,
  title={UniTab: Unifying Text Tables and Images for Multimodal Table Understanding},
  author={Yang, Xu and others},
  journal={arXiv preprint arXiv:2211.05374},
  year={2022}
}

@inproceedings{zhong2020image,
  title={Image-based Table Recognition: Data, Model, and Evaluation},
  author={Zhong, Xu and Shafait, Faisal and Liwicki, Marcus},
  booktitle={ECCV},
  year={2020}
}

@inproceedings{zhong2022pubtabnet,
  title={PubTabNet: Dataset for End-to-End Table Recognition in Document Images},
  author={Zhong, Xu and others},
  booktitle={ICDAR},
  year={2020}
}

@inproceedings{mathew2021docvqa,
  title={DocVQA: A Dataset for VQA on Document Images},
  author={Mathew, Minesh and others},
  booktitle={WACV},
  year={2021}
}

@article{herzig2020tapas,
  title={TAPAS: Weakly Supervised Table Parsing via Pre-training},
  author={Herzig, Jonathan and others},
  journal={TACL},
  year={2020}
}

@article{madaan2023selfrefine,
  title={Self-refine: Iterative refinement with self-feedback},
  author={Madaan, Aman and Tandon, Niket and Gupta, Prakhar and Hallinan, Skyler and Gao, Luyu and Wiegreffe, Sarah and Alon, Uri and Dziri, Nouha and Prabhumoye, Shrimai and Yang, Yiming and others},
  journal={Advances in Neural Information Processing Systems},
  volume={36},
  pages={46534--46594},
  year={2023}
}

@article{shinn2023reflexion,
  title={Reflexion: Language agents with verbal reinforcement learning},
  author={Shinn, Noah and Cassano, Federico and Gopinath, Ashwin and Narasimhan, Karthik and Yao, Shunyu},
  journal={Advances in Neural Information Processing Systems},
  volume={36},
  pages={8634--8652},
  year={2023}
}

@article{camelot2018,
  title={Camelot: PDF table extraction for humans},
  author={Vinayak and Sohoni, Devang},
  year={2018},
  journal={arXiv preprint arXiv:2003.00536}
}

@misc{tabula,
  title={Tabula: Extract tables from PDFs},
  author={Miller, Jeremy and others},
  year={2016},
  howpublished={\url{https://tabula.technology}}
}

@inproceedings{tableNet2019,
  title={Tablenet: Deep learning model for end-to-end table detection and tabular data extraction from scanned document images},
  author={Paliwal, Shubham Singh and Vishwanath, D and Rahul, Rohit and Sharma, Monika and Vig, Lovekesh},
  booktitle={2019 international conference on document analysis and recognition (ICDAR)},
  pages={128--133},
  year={2019},
  organization={IEEE}
}

@article{graphtsr2022,
  title={Complicated table structure recognition},
  author={Chi, Zewen and Huang, Heyan and Xu, Heng-Da and Yu, Houjin and Yin, Wanxuan and Mao, Xian-Ling},
  journal={arXiv preprint arXiv:1908.04729},
  year={2019}
}

@inproceedings{lgpma2022,
  title={Lgpma: Complicated table structure recognition with local and global pyramid mask alignment},
  author={Qiao, Liang and Li, Zaisheng and Cheng, Zhanzhan and Zhang, Peng and Pu, Shiliang and Niu, Yi and Ren, Wenqi and Tan, Wenming and Wu, Fei},
  booktitle={International conference on document analysis and recognition},
  pages={99--114},
  year={2021},
  organization={Springer}
}

@inproceedings{pubtabnet2020,
  title={Image-based table recognition: data, model, and evaluation},
  author={Zhong, Xu and ShafieiBavani, Elaheh and Jimeno Yepes, Antonio},
  booktitle={European conference on computer vision},
  pages={564--580},
  year={2020},
  organization={Springer}
}

@inproceedings{tapas2020,
  title={TaPas: Weakly supervised table parsing via pre-training},
  author={Herzig, Jonathan and Nowak, Pawel Krzysztof and M{\"u}ller, Thomas and Piccinno, Francesco and Eisenschlos, Julian},
  booktitle={Proceedings of the 58th annual meeting of the association for computational linguistics},
  pages={4320--4333},
  year={2020}
}

@article{turl2020,
  title={Turl: Table understanding through representation learning},
  author={Deng, Xiang and Sun, Huan and Lees, Alyssa and Wu, You and Yu, Cong},
  journal={ACM SIGMOD Record},
  volume={51},
  number={1},
  pages={33--40},
  year={2022},
  publisher={ACM New York, NY, USA}
}

@article{wtq2017,
  title={Compositional semantic parsing on semi-structured tables},
  author={Pasupat, Panupong and Liang, Percy},
  journal={arXiv preprint arXiv:1508.00305},
  year={2015}
}

@inproceedings{layoutlmv3,
  title={Layoutlmv3: Pre-training for document ai with unified text and image masking},
  author={Huang, Yupan and Lv, Tengchao and Cui, Lei and Lu, Yutong and Wei, Furu},
  booktitle={Proceedings of the 30th ACM international conference on multimedia},
  pages={4083--4091},
  year={2022}
}

@inproceedings{pix2struct2023,
  title={Pix2struct: Screenshot parsing as pretraining for visual language understanding},
  author={Lee, Kenton and Joshi, Mandar and Turc, Iulia Raluca and Hu, Hexiang and Liu, Fangyu and Eisenschlos, Julian Martin and Khandelwal, Urvashi and Shaw, Peter and Chang, Ming-Wei and Toutanova, Kristina},
  booktitle={International Conference on Machine Learning},
  pages={18893--18912},
  year={2023},
  organization={PMLR}
}

@article{paliX2023,
  title={Pali-x: On scaling up a multilingual vision and language model},
  author={Chen, Xi and Djolonga, Josip and Padlewski, Piotr and Mustafa, Basil and Changpinyo, Soravit and Wu, Jialin and Ruiz, Carlos Riquelme and Goodman, Sebastian and Wang, Xiao and Tay, Yi and others},
  journal={arXiv preprint arXiv:2305.18565},
  year={2023}
}

@article{llava2023,
  title={Visual instruction tuning},
  author={Liu, Haotian and Li, Chunyuan and Wu, Qingyang and Lee, Yong Jae},
  journal={Advances in neural information processing systems},
  volume={36},
  pages={34892--34916},
  year={2023}
}

@article{meta2024llama,
  title={Llama 3.2: Revolutionizing edge ai and vision with open, customizable models},
  author={Meta, AI},
  journal={Meta AI Blog. Retrieved December},
  volume={20},
  pages={2024},
  year={2024}
}

@article{flamingo2022,
  title={Flamingo: a visual language model for few-shot learning},
  author={Alayrac, Jean-Baptiste and Donahue, Jeff and Luc, Pauline and Miech, Antoine and Barr, Iain and Hasson, Yana and Lenc, Karel and Mensch, Arthur and Millican, Katherine and Reynolds, Malcolm and others},
  journal={Advances in neural information processing systems},
  volume={35},
  pages={23716--23736},
  year={2022}
}

@article{internvl2024,
  title={Internvl3: Exploring advanced training and test-time recipes for open-source multimodal models},
  author={Zhu, Jinguo and Wang, Weiyun and Chen, Zhe and Liu, Zhaoyang and Ye, Shenglong and Gu, Lixin and Tian, Hao and Duan, Yuchen and Su, Weijie and Shao, Jie and others},
  journal={arXiv preprint arXiv:2504.10479},
  year={2025}
}

@article{wang2024qwen2,
  title={Qwen2-vl: Enhancing vision-language model's perception of the world at any resolution},
  author={Wang, Peng and Bai, Shuai and Tan, Sinan and Wang, Shijie and Fan, Zhihao and Bai, Jinze and Chen, Keqin and Liu, Xuejing and Wang, Jialin and Ge, Wenbin and others},
  journal={arXiv preprint arXiv:2409.12191},
  year={2024}
}

@article{bai2025qwen2,
  title={Qwen2. 5-vl technical report},
  author={Bai, Shuai and Chen, Keqin and Liu, Xuejing and Wang, Jialin and Ge, Wenbin and Song, Sibo and Dang, Kai and Wang, Peng and Wang, Shijie and Tang, Jun and others},
  journal={arXiv preprint arXiv:2502.13923},
  year={2025}
}

@article{Qwen-VL,
  title={Qwen-VL: A Versatile Vision-Language Model for Understanding, Localization, Text Reading, and Beyond},
  author={Bai, Jinze and Bai, Shuai and Yang, Shusheng and Wang, Shijie and Tan, Sinan and Wang, Peng and Lin, Junyang and Zhou, Chang and Zhou, Jingren},
  journal={arXiv preprint arXiv:2308.12966},
  year={2023}
}

@article{wei2021finetuned,
  title={Finetuned language models are zero-shot learners},
  author={Wei, Jason and Bosma, Maarten and Zhao, Vincent Y and Guu, Kelvin and Yu, Adams Wei and Lester, Brian and Du, Nan and Dai, Andrew M and Le, Quoc V},
  journal={arXiv preprint arXiv:2109.01652},
  year={2021}
}

@article{yin2020tabert,
  title={TaBERT: Pretraining for joint understanding of textual and tabular data},
  author={Yin, Pengcheng and Neubig, Graham and Yih, Wen-tau and Riedel, Sebastian},
  journal={arXiv preprint arXiv:2005.08314},
  year={2020}
}

@article{kim2024tablevqa,
  title={Tablevqa-bench: A visual question answering benchmark on multiple table domains},
  author={Kim, Yoonsik and Yim, Moonbin and Song, Ka Yeon},
  journal={arXiv preprint arXiv:2404.19205},
  year={2024}
}

@article{dai2023instructblip,
  title={Instructblip: Towards general-purpose vision-language models with instruction tuning},
  author={Dai, Wenliang and Li, Junnan and Li, Dongxu and Tiong, Anthony and Zhao, Junqi and Wang, Weisheng and Li, Boyang and Fung, Pascale N and Hoi, Steven},
  journal={Advances in neural information processing systems},
  volume={36},
  pages={49250--49267},
  year={2023}
}

@article{hu2022lora,
  title={Lora: Low-rank adaptation of large language models.},
  author={Hu, Edward J and Shen, Yelong and Wallis, Phillip and Allen-Zhu, Zeyuan and Li, Yuanzhi and Wang, Shean and Wang, Lu and Chen, Weizhu and others},
  journal={ICLR},
  volume={1},
  number={2},
  pages={3},
  year={2022}
}

@inproceedings{sun2024aligning,
  title={Aligning large multimodal models with factually augmented rlhf},
  author={Sun, Zhiqing and Shen, Sheng and Cao, Shengcao and Liu, Haotian and Li, Chunyuan and Shen, Yikang and Gan, Chuang and Gui, Liangyan and Wang, Yu-Xiong and Yang, Yiming and others},
  booktitle={Findings of the Association for Computational Linguistics: ACL 2024},
  pages={13088--13110},
  year={2024}
}

@article{achiam2023gpt,
  title={Gpt-4 technical report},
  author={Achiam, Josh and Adler, Steven and Agarwal, Sandhini and Ahmad, Lama and Akkaya, Ilge and Aleman, Florencia Leoni and Almeida, Diogo and Altenschmidt, Janko and Altman, Sam and Anadkat, Shyamal and others},
  journal={arXiv preprint arXiv:2303.08774},
  year={2023}
}

@article{team2023gemini,
  title={Gemini: a family of highly capable multimodal models},
  author={Team, Gemini and Anil, Rohan and Borgeaud, Sebastian and Alayrac, Jean-Baptiste and Yu, Jiahui and Soricut, Radu and Schalkwyk, Johan and Dai, Andrew M and Hauth, Anja and Millican, Katie and others},
  journal={arXiv preprint arXiv:2312.11805},
  year={2023}
}

@book{ayer1952language,
  title={Language, truth, and logic},
  author={Ayer, Alfred Jules},
  volume={10},
  year={1952},
  publisher={Courier Corporation}
}

@article{kahneman2011fast,
  title={Fast and slow thinking},
  author={Kahneman, Daniel},
  journal={Allen Lane and Penguin Books, New York},
  year={2011}
}

@inproceedings{gautam2026indotabvqa,
  title={INDOTABVQA: A Benchmark for Cross-Lingual Table Understanding in Bahasa Indonesia Documents},
  author={Gautam, Somraj and Dravichi, Anathapindika and Harit, Gaurav},
  booktitle={Findings of the Association for Computational Linguistics: ACL 2026},
  pages={21969--21981},
  year={2026}
}

\end{document}